\documentclass{article}
\usepackage{spconf,amsmath,graphicx,url}

\title{END-TO-END ASR-FREE KEYWORD SEARCH FROM SPEECH}
\name{Kartik Audhkhasi\thanks{The authors thank Dogan Can and Shrikanth Narayanan of the Signal Analysis and Interpretation Lab, University of Southern California for useful discussions. This paper uses the IARPA-Babel404b-v1.0a full language pack. Supported by the Intelligence Advanced Research Projects Activity (IARPA) via Department of Defense U.S. Army Research Laboratory (DoD/ARL) contract number W911NF-12-C-0012. The U.S. Government is authorized to reproduce and distribute reprints for Governmental purposes notwithstanding any copyright annotation thereon.
}, Andrew Rosenberg, Abhinav Sethy, Bhuvana Ramabhadran, Brian Kingsbury}
\address{IBM Watson, IBM T. J. Watson Research Center, Yorktown Heights, New York
}

\begin{document}
\newpage
{\onecolumn
Copyright 2017 IEEE. Published in the IEEE 2017 International Conference on Acoustics, Speech, and Signal Processing (ICASSP 2017), scheduled for 5-9 March 2017 in New Orleans, Louisiana, USA. Personal use of this material is permitted. However, permission to reprint/republish this material for advertising or promotional purposes or for creating new collective works for resale or redistribution to servers or lists, or to reuse any copyrighted component of this work in other works, must be obtained from the IEEE. Contact: Manager, Copyrights and Permissions / IEEE Service Center / 445 Hoes Lane / P.O. Box 1331 / Piscataway, NJ 08855-1331, USA. Telephone: + Intl. 908-562-3966.}
\twocolumn
\ninept
\maketitle
\begin{abstract}
End-to-end (E2E) systems have achieved competitive results compared to conventional hybrid hidden Markov model (HMM)-deep neural network based automatic speech recognition (ASR) systems. Such E2E systems are attractive due to the lack of dependence on alignments between input acoustic and output grapheme or HMM state sequence during training. This paper explores the design of an ASR-free end-to-end system for text query-based keyword search (KWS) from speech trained with minimal supervision. Our E2E KWS system consists of three sub-systems. The first sub-system is a recurrent neural network (RNN)-based acoustic auto-encoder trained to reconstruct the audio through a finite-dimensional representation. The second sub-system is a character-level RNN language model using embeddings learned from a convolutional neural network. Since the acoustic and text query embeddings occupy different representation spaces, they are input to a third feed-forward neural network that predicts whether the query occurs in the acoustic utterance or not. This E2E ASR-free KWS system performs respectably despite lacking a conventional ASR system and trains much faster.
\end{abstract}
\begin{keywords}
End-to-end systems, neural networks, keyword search, automatic speech recognition
\end{keywords}
\section{Introduction}
\label{sec:intro}
Deep neural networks (DNNs) have pushed the state-of-the-art for automatic speech recognition (ASR) systems~\cite{hinton2012deep}. This has led to significant performance improvements on several well-known ASR benchmarks such as Switchboard~\cite{saon2016ibm,xiong2016ms}. End-to-end (E2E) or fully-neural architectures have become an alternative to the hybrid hidden Markov model (HMM)-DNN architecture. These include the connectionist temporal classification~\cite{graves2006connectionist,miao2015eesen} loss-based recurrent neural network (RNN) and attention-based RNNs~\cite{bahdanau2014neural,bahdanau2016end}.  

Automatic speech recognition is often not the end goal of real-world speech information processing systems. Instead, an important end goal is information retrieval, in particular keyword search (KWS), that involves retrieving the speech utterances containing a user-specified text query from a large database. Conventional KWS from speech uses an ASR system as a front-end that converts the speech database into a finite-state transducer (FST) index containing all possible hypotheses word sequences with their associated confidence scores and time stamps~\cite{can2011lattice}. The user-specific text query is then composed with this FST index to find putative keyword locations and confidence scores.

Training a good ASR system is time-consuming and requires substantial amount of transcribed audio data. The main novelty of this paper is an end-to-end ASR-free KWS system motivated by the recent success of E2E systems for ASR. Our fully-neural E2E KWS system lacks both an ASR system and a FST-index, which makes it fast to train. We train our system in only 2 hours with minimal supervision and do not need fully transcribed training audio. 
Our system performs significantly better than chance and respectably compared to a state-of-the-art hybrid HMM-DNN ASR-based system which takes over 72 hours to train.

The next section gives an overview of related prior work. Section~\ref{sec:e2e_system} introduces our E2E ASR-free system by discussing its three constituent sub-systems - an RNN-based acoustic encoder-decoder, a convolutional neural network (CNN)-RNN character language model (LM), and a feed-forward KWS network. Section~\ref{sec:expts} discusses the experimental setup, training of the E2E KWS system, and analysis of the results. Section~\ref{sec:concl} gives directions for future work.

\vspace{-10pt}
\subsection{Prior Work}
\label{sec:prior}
Most prior works relevant to this paper have focused on query-by-example (QbyE) retrieval of speech from a database. The user provides a speech utterance of the query to be searched, in contrast to a text query used in the KWS setup of this paper. Dynamic time warping (DTW) of acoustic features extracted from the speech query and speech utterances from the database is a classic technique in such QbyE systems~\cite{hazen2009query,zhang2009unsupervised}. 
The cost of DTW alignment serves as a matching score for retrieval. 

Chen, Parada, and Sainath~\cite{chen2015query} present a system for QbyE where the audio database contains examples of certain key-phrases, such as \emph{``hello genie''}. The last $k$ state vectors from the final hidden layer of an RNN acoustic model give a fixed-dimensional representation for both the speech query and each utterance in the speech database. The KWS detector then computes cosine similarity between the speech query and utterance representations. Levin et. al~\cite{levin2013fixed} compares several non-neural network-based techniques for computing fixed-dimensional representations of speech segments for QbyE, including principal component analysis and Laplacian eigenmaps.
Chung et. al~\cite{chung2016audio} also present a similar QbyE system using an acoustic RNN encoder-decoder network. Kamper, Wang, and Livescu~\cite{kamper2016deep} use a Siamese convolutional neural network (CNN) for obtaining acoustic embeddings of the audio query and utterances from the database, and train this network by minimizing a triplet hinge loss.

Our paper is closely related to the recent work of Palaz, Synnaeve, and Collobert~\cite{palaz2016jointly}, where the authors propose a CNN-based ASR system trained on a bag of words in the speech utterance. The CNN emits a time sequence of posterior distributions over the vocabulary, which is then aggregated over time to produce an estimate of the bag of words. However, we note that their system is trained on a fixed vocabulary of 1000 words compared with our proposed open-vocabulary system that does not use word identity. In addition, the training examples used in their approach use a stronger supervision of word identity compared with a much weaker supervision in our proposed model. The next section presents the architecture of our E2E ASR-free KWS system.

\section{End-to-End ASR-free KWS Architecture}
\label{sec:e2e_system}
Our E2E ASR-free KWS system is philosophically similar to a conventional hybrid ASR-based KWS system, with three sub-systems that model the acoustics, language, and keyword search. However, there are several differences in the structures of these sub-systems and their training.

\subsection{RNN Acoustic Auto-encoder}
\begin{figure}[h]

\begin{center}
\includegraphics[scale=0.26]{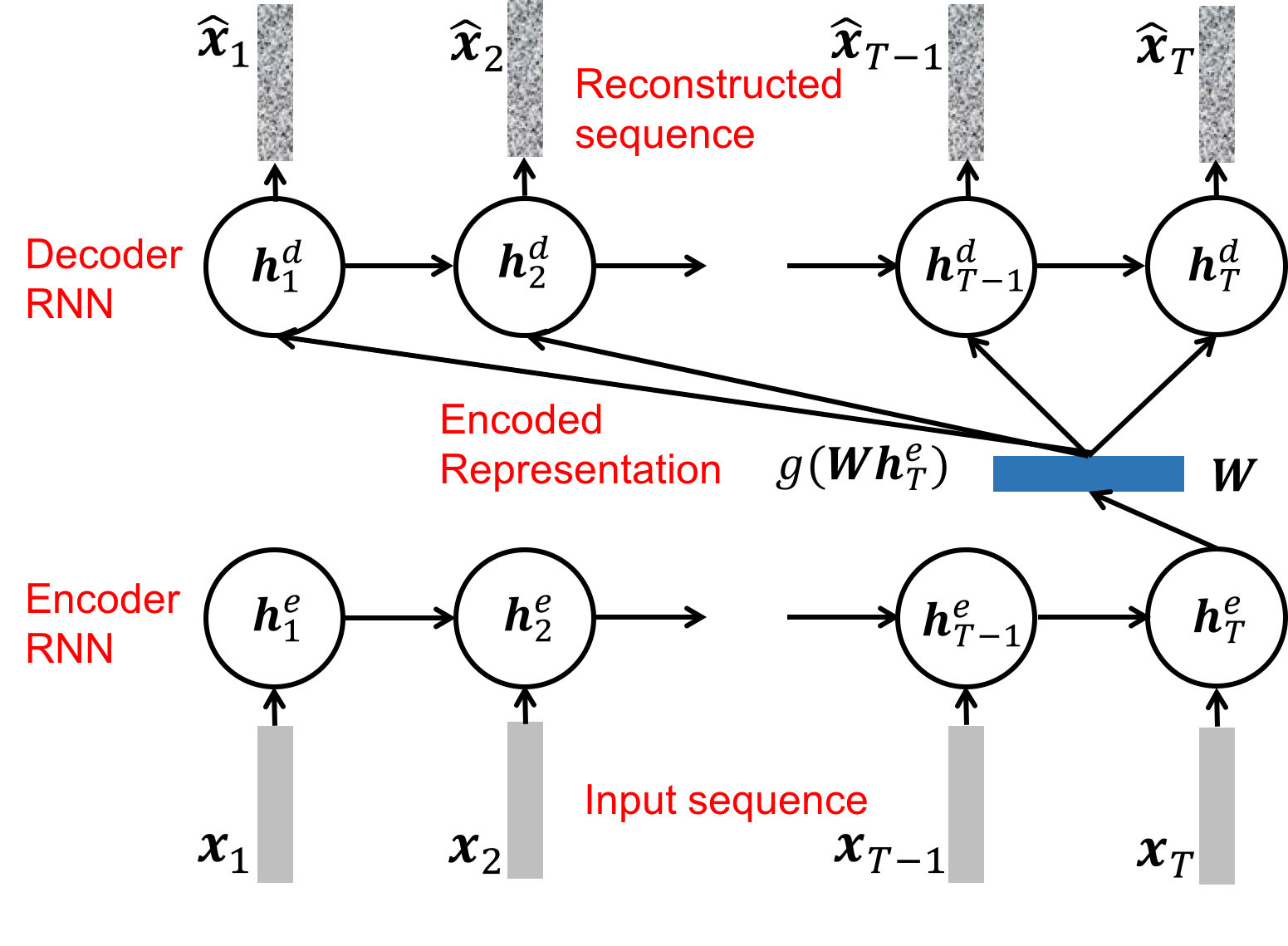}
\caption{This figure shows an RNN acoustic auto-encoder for a $T$-length input sequence of acoustic feature vectors through a $D$-dimensional encoded representation $g(\mathbf{W}\mathbf{h}_T^e)$, where $g$ denotes a ReLU activation function.}
\label{fig:rnn_autoenc}
\end{center}
\end{figure}

Motivated by prior work~\cite{chen2015query,chung2016audio} on computing fixed-dimensional representations from variable length acoustic feature vector sequence, we use an RNN-based auto-encoder as shown in Figure~\ref{fig:rnn_autoenc}. The encoder processes $T$ acoustic feature vectors $(\mathbf{x}_1,\ldots,\mathbf{x}_T)$ by a uni-directional RNN with gated recurrent unit (GRU)~\cite{chung2014empirical} hidden units unrolled over $T$ time steps. A fully-connected layer with weight matrix $\mathbf{W}$ and $D$ rectified linear units (ReLUs) denoted by $g$ then processes the
hidden state vector $\mathbf{h}^e_T$ from time step $T$. A decoder GRU-RNN takes the resulting $D$-dimensional acoustic representation $g(\mathbf{Wh}^e_T)$ as input at each time step $t \in \{1,\ldots,T\}$ to reconstruct the original sequence of $T$ acoustic feature vectors.

\vspace{-15pt}
We use the $D$-dimensional output $g(\mathbf{Wh}^e_T)$ of the RNN encoder as the vector representation of the input acoustic feature vector sequence. This ``acoustic model'' does not use any transcribed speech data to train, in contrast with an acoustic model in a conventional ASR system. This is in line with our overall goal of making the entire KWS system ASR-free and train with less supervision. The next section describes the language model used to produce a fixed dimensional representation of the text query.

\vspace{-10pt}
\subsection{CNN-RNN Character LM}
\begin{figure}[h]
\begin{center}
\includegraphics[scale=0.30]{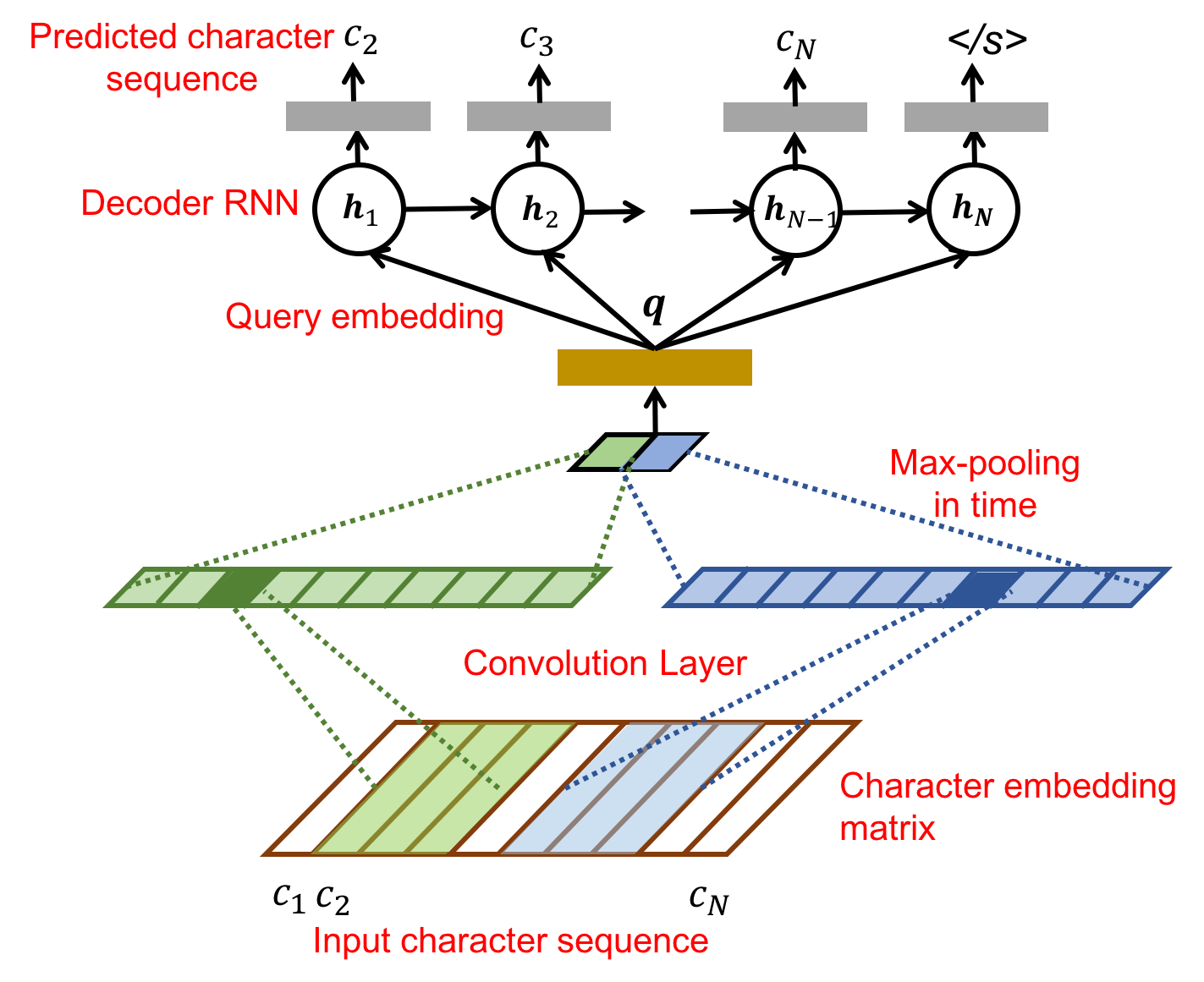}
\caption{This figure shows an character CNN-RNN LM for encoding text queries. We show two convolutional masks for simplicity.}
\label{fig:char_rnnlm}
\end{center}
\end{figure}

We use the CNN-based character RNN-LM architecture from Kim et. al~\cite{kim2015character} for deriving query embeddings. Figure~\ref{fig:char_rnnlm} shows this LM for the simple case of 2 convolutional masks. We map the input sequence of $N$ characters $(c_1,\ldots,c_N)$ to a matrix of $d$-dimensional character embeddings via a look-up table. Next, $M$ $d\times w$ convolutional masks operate on the resulting $d \times N$ embedding matrix to produce a set of $M$ $N$-dimensional vectors, one per mask. We then perform max-pooling over time on each of these vectors to obtain a scalar per mask and a $M$-dimensional embedding vector. This embedding vector then feeds into a GRU-RNN that predicts one out of $K$ characters at each time step.

\vspace{-15pt}
One key difference between our LM and the one from Kim et. al~\cite{kim2015character} is that we train our LM to predict a sequence of characters instead of words. 
The next section presents our overall KWS system that uses learned acoustic and query embeddings to predict whether the query occurs in the utterance or not.

\begin{figure*}[h]
\begin{center}
\includegraphics[scale=0.38]{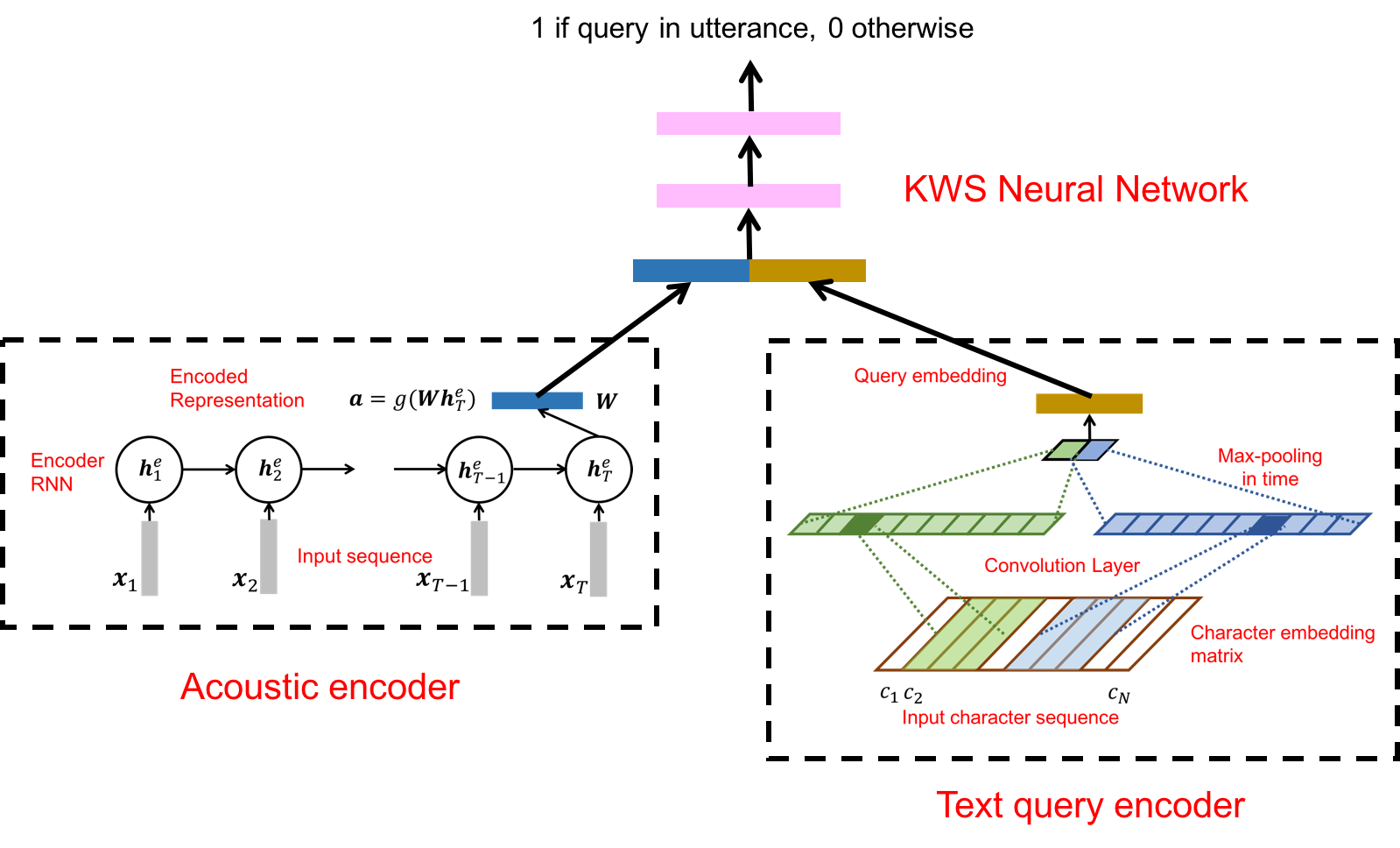}
\caption{This figure shows the overall E2E KWS system. The finite-dimensional embeddings from the acoustic and query encoders feed into the KWS neural network that predict if the query occurs in the utterance or not.}
\label{fig:e2e_kws_overall}
\end{center}
\end{figure*}
\vspace{-10pt}
\subsection{Overall E2E ASR-free KWS System}
The final block in the overall KWS system is a neural network that takes the speech utterance and text query embeddings as input and predicts whether the query occurs in the utterance or not. This in contrast to previous works on QbyE in speech where both the speech utterance and speech query lie in the same acoustic representation space and cosine similarity is enough to match the two. Figure~\ref{fig:e2e_kws_overall} shows the overall E2E KWS system. We extract the encoders from both the acoustic RNN auto-encoder and the CNN-RNN character LM, and feed them into a feed-forward neural network. In contrast to conventional ASR-based approaches to KWS from speech, the E2E system in Figure~\ref{fig:e2e_kws_overall} is also jointly trainable after the utterance and query encoders have been pre-trained. The next section presents our data preparation, experiments, results and analysis.

\section{Experiments, Results, and Analysis}
\label{sec:expts}
\subsection{Data Description and Preparation}
We used the Georgian full-language pack from Option Period 3 of the IARPA Babel program for keyword search from low resource languages. The training data contains 40 hours of transcribed audio corresponding to 45k utterances and a graphemic pronunciation dictionary containing 35k words. We use the 15 hour development audio, 2k in-vocabulary (IV) keywords and 400 out-of-vocabulary (OOV) keywords for testing. The keywords include both single-word and multi-word queries. Multilingual acoustic features have been successful in the Babel program compared to conventional features such as Mel frequency cepstral coefficients. We use an 80-dimensional multilingual acoustic front-end~\cite{cui2015multilingual} trained on all 24 Babel languages from the Base Period to Option Period 3, excluding Georgian.

The training of the acoustic auto-encoder RNN and CNN-RNN character LM does not requires special selection of training examples. However, the final KWS neural network requires a set of positive and negative examples to train. We wanted to keep the training of this network independent of the list of test queries. Hence we constructed positive examples by taking all words in the 35k vocabulary and finding a maximum of 100 utterances that contain each word. We then selected an equal number of negative examples by picking utterances that do not contain the particular vocabulary word. We ensured good coverage of the acoustic training data by constraining each acoustic utterance to be picked in only a maximum of 5 positive and negative examples. We also excluded acoustic utterances that did not contain speech. This resulted in 62k positive and 62k negative examples each for training the KWS network. We processed the development audio and test queries in similar fashion, and ended up with 7.5k positive and negative examples each for testing. We implemented the system in Keras~\cite{chollet2015keras} and Theano~\cite{2016arXiv160502688short}. 

\subsection{Acoustic Auto-encoder}
\begin{figure}[h]
\begin{center}
\includegraphics[scale=0.37]{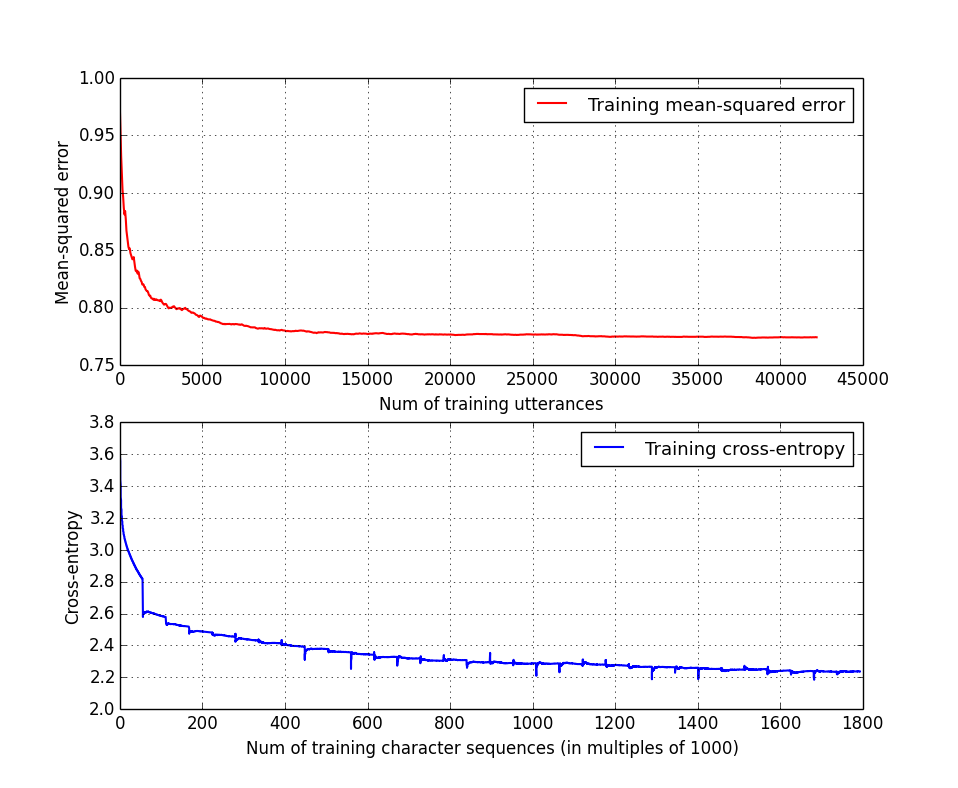}
\caption{This figure shows the training loss of the RNN acoustic encoder-decoder (top plot) and the CNN-RNN character LM (bottom plot) as training proceeds.}
\label{fig:training_loss_encoders}
\end{center}
\end{figure}

We used 300 hidden GRU neurons in the acoustic encoder and decoder RNNs. We sorted 
the acoustic feature sequences in the training data set in increasing order of length since it improved training convergence. We unrolled both the RNNs for 15 seconds or 1500 time steps which is the length of the maximum acoustic utterance in the training data set. We padded all acoustic feature sequences to make their length equal to 1500 time steps and excluded these extra frames from the loss function and gradient computation. We used a linear dense layer of size $300$ to compute the embedding from the final hidden state vector of the encoder RNN. We trained the acoustic auto-encoder by minimizing the mean-squared reconstruction error of the input sequence of acoustic feature vectors using the Adam optimization algorithm~\cite{kingma2014adam} with a mini-batch size of 40 utterances and learning rate of $1\times 10^{-3}$. We used the ``newbob'' annealing schedule by reducing the learning rate by half whenever the validation set loss did not decrease sufficiently. The top plot of Figure~\ref{fig:training_loss_encoders} shows the progress of the training set loss as training proceeds. We find that the loss drops significantly in the initial part of training and is nearly constant after 15k utterances. Further epochs through the training data did not yield significant improvements in loss. The next section discusses details about the training of the CNN-RNN character LM.

\begin{table*}[h]
\caption{This table compares the KWS accuracy of the E2W KWS and DNN-HMM hybrid ASR systems for different IV query lengths.}
\begin{center}
\begin{tabular}{|c|c|c|c|c|c|c|c|c|c|c|c|c|c|}
\hline
\bf{Query Length} $\rightarrow$ & $\leq$3 & 4 & 5 & 6 & 7 & 8 & 9 & 10 & 11 & 12 & 13 & 14 & $\geq$15 \\ \hline
DNN-HMM (2gm word LM) & 69.8 & 72.5 & 74.6 & 77.9 & 77.3 & 78.8 & 76.7 & 80.0 & 78.7 & 78.9 & 74.5 & 77.1 & 78.6 \\ \hline
DNN-HMM (4gm grapheme LM) & 70.6 & 74.7 & 71.1 & 72.8 & 71.9 & 70.1 & 66.4 & 68.4 & 67.3 & 65.2 & 65.6 & 65.4 & 65.3 \\ \hline
E2E ASR-free & 51.8 & 56.4 & 56.5 & 55.6 & 55.3 & 55.1 & 55.7 & 52.1 & 53.5 & 58.4 & 55.8 & 56.7 & 60.0 \\ \hline
\end{tabular}
\end{center}
\label{tab:class_perf}
\end{table*}%
\vspace{-10pt}

\begin{table}[h]
\caption{This table compares the KWS accuracy of the E2W KWS and DNN-HMM hybrid ASR systems for IV and OOV queries.}
\vspace{-10pt}
\begin{center}
\begin{tabular}{|c|c|c|}
\hline
\bf{Query Type} $\rightarrow$ & \bf{IV} & \bf{OOV} \\ \hline
DNN-HMM (2gm word LM) & 76.7 & 50.0 (chance) \\ \hline
DNN-HMM (4gm grapheme LM) & 70.7 & 55.5 \\ \hline
E2E ASR-free & 55.6 & 57.7 \\ \hline
\end{tabular}
\end{center}
\label{tab:class_perf_iv_oov}
\end{table}%

\vspace{-10pt}

\begin{figure}[h]
\begin{center}
\includegraphics[scale=0.40]{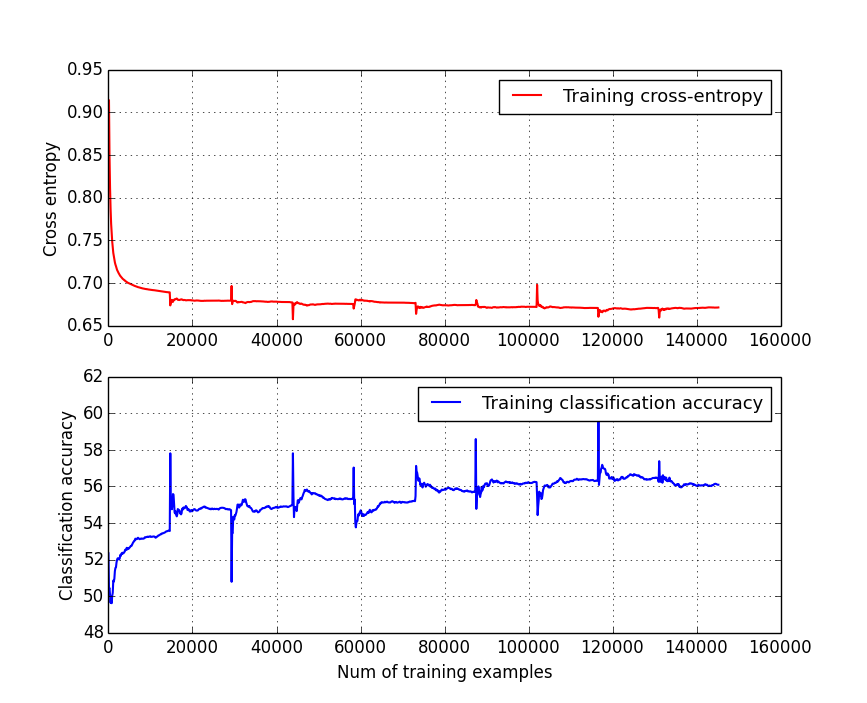}
\caption{This figure shows the training loss and classification accuracy of the KWS neural network as training proceeds.}
\label{fig:training_loss_kwsnet}
\end{center}
\end{figure}


\vspace{-10pt}
\subsection{CNN-RNN Character LM}
We used all the acoustic transcripts and converted them to sentences over 39 unique graphemes. We broke sentences longer than 50 graphemes into smaller chunks for preparing mini-batches, since the maximum length of a query is 23 graphemes. We used 50-dimensional embeddings for each grapheme and 300 convolutional masks of size $50 \times 3$, which resulted in a 300-dimensional embedding for each input sequence. The decoder RNN used 256 GRUs and a softmax layer of 39 neurons at each time step. We minimized the cross-entropy of the output grapheme sequence, and used Adam with a mini-batch size of 256 sequences, learning rate of $1\times 10^{-3}$, and the newbob annealing schedule. The bottom plot in Figure~\ref{fig:training_loss_encoders} shows the progress of training cross-entropy.Unlike the acoustic RNN auto-encoder, we trained this network for a few epochs.


\subsection{KWS Neural Network}
After training the acoustic RNN auto-encoder and CNN-RNN character LM, we removed the decoders from both models, concatenated the encoder outputs (resulting in a 600-dimensional vector), and fed them into a fully-connected feed-forward neural network with one ReLU hidden layer of size 256. The output layer contained two softmax neurons that detect whether the input query occurred in the acoustic utterance or not. We applied 50\% dropout to all layers of this network since it improved classification performance. We used Adam to train this network by minimizing the cross-entropy loss with a batch size of 128 examples, learning rate of $1 \times 10^{-3}$, and the newbob learning rate annealing schedule. We first back-propagated the errors through this KWS network only for a few epochs, and then through the entire network.The latter did not have a significant impact on KWS accuracy. 
Figure~\ref{fig:training_loss_kwsnet} shows the training loss and classification accuracy of this KWS neural network over several epochs. We observe that the network gradually reaches an above-chance classification accuracy of approximately 56\%.

We then tested the KWS network on the test set of 2k in-vocabulary (IV), 400 out-of-vocabulary (OOV) queries and 15 hours of development audio. To get the topline performance, we also trained a hybrid DNN-HMM system for Georgian. This ASR system used 6000 context-dependent states in the HMM and a five-layer deep neural network with 1024 neurons in each layer. We trained this network first using the frame-wise cross-entropy criterion and then using Hessian-free sequence minimum Bayes risk (sMBR) training~\cite{martens2010deep,kingsbury2012scalable}. We used two LMs - a bigram word LM trained on a vocabulary of 35k words and a 4-gram grapheme LM trained over 39 graphemes. The word error rate (WER) of this hybrid ASR system is 41.9\%. We then performed KWS over the 1-best transcript obtained by Viterbi decoding of the development audio, instead of the full-blown lattice-based KWS for simplicity and a fair-comparison to the E2E KWS approach.

Table~\ref{tab:class_perf_iv_oov} shows the classification accuracies of the DNN-HMM ASR system and the proposed E2E ASR-free KWS system. We obtain a classification accuracy of 55.6\% on IV and 57.7\% on OOV queries, which is significantly above chance. As expected, the IV performance is lower than that of the hybrid ASR system using 2-gm word LM. But it is interesting to note that the E2E ASR-free and hybrid system using 4-gm grapheme LM have closer accuracies, especially for OOV queries, where the E2E KWS system performs better by 2.2\% absolute. This result is encouraging, since the hybrid system uses word-level transcriptions for training the acoustic model and 36 times more training time than the E2E ASR-free KWS system. We performed further analysis of the dependence of KWS performance on query length. Table~\ref{tab:class_perf} shows the classification accuracy as a function of number of graphemes in the query. We observe that both the ASR-based and E2E KWS systems have difficulty detecting short queries. In case of the E2E system, this is because it is difficult to derive a reliable representation for short queries due to the lack of context. A key advantage of the E2E KWS system is that it takes 36 times less time to train than the DNN-HMM system.

\section{Conclusions and Future Work}
\label{sec:concl}
This paper presented a novel end-to-end ASR-free approach to text query-based KWS from speech. This is in contrast to ASR-based approaches and to previous works on query-by-example retrieval of audio. The proposed system trains with minimal supervision without any transcription of the acoustic data. The system uses an RNN acoustic auto-encoder, a CNN-RNN character LM, and a KWS neural network that decides whether the input text query occurs in the acoustic utterance. We show that the system performs respectably on a Georgian keyword search task from the Babel program, and trains 36 times faster than a conventional DNN-HMM hybrid ASR system. Future work should focus on closing the performance gap with the hybrid ASR system and estimating times of the detected keywords.


\bibliographystyle{IEEEbib}
\bibliography{e2e_kws_refs}

\end{document}